\pdfoutput=1

\documentclass[11pt]{article}

\usepackage[]{acl}

\usepackage{times}
\usepackage{latexsym}

\usepackage[T1]{fontenc}

\usepackage[utf8]{inputenc}

\usepackage{microtype}

\usepackage{hyperref}
\usepackage{url}
\usepackage{kotex}
\usepackage{caption}
\usepackage{multirow}
\usepackage[flushleft]{threeparttable}
\usepackage{makecell}
\usepackage{tabulary}
\usepackage{tabularx}
\usepackage{graphicx}
\usepackage{wrapfig}
\usepackage{amsmath}
\usepackage{amssymb}
\usepackage{tcolorbox}
\tcbuselibrary{breakable}
\usepackage{amsmath}

\DeclareMathOperator*{\argmin}{arg\,min}

\usepackage{xspace}
\newcommand{\eg}{\textit{e.g.}\xspace}

\title{Toward Interactive Regional Understanding in \\Vision-Large Language Models}

\makeatletter
\def\thanks#1{\protected@xdef\@thanks{\@thanks
        \protect\footnotetext{#1}}}
\makeatother

\author{Jungbeom Lee$^{1*}$\thanks{$^*$ Work done while doing visiting researcher at NAVER} ~~~~~~~~~~~~ Sanghyuk Chun$^{2\dagger}$ ~~~~~~~~~~~~ Sangdoo Yun$^{2\dagger}$\thanks{$\dagger$ Corresponding authors}
\\
$^1$Amazon ~~~~~~~~~~
$^2$NAVER AI Lab \\
{\tt\small jungbeol@amazon.com, \{sanghyuk.c, sangdoo.yun\}@navercorp.com}
}

\begin{document}
\maketitle
\begin{abstract}
Recent Vision-Language Pre-training (VLP) models have demonstrated significant advancements. Nevertheless, these models heavily rely on image-text pairs that capture only coarse and global information of an image, leading to a limitation in their regional understanding ability. In this work, 
we introduce \textbf{RegionVLM}, equipped with explicit regional modeling capabilities, allowing them to understand user-indicated image regions. To achieve this, we design a simple yet innovative architecture, requiring no modifications to the model architecture or objective function. Additionally, we leverage a dataset that contains a novel source of information, namely Localized Narratives, which has been overlooked in previous VLP research. Our experiments demonstrate that our single generalist model not only achieves an interactive dialogue system but also exhibits superior performance on various zero-shot region understanding tasks, without compromising its ability for global image understanding.

\end{abstract}

\section{Introduction}

Vision-Language Pre-training (VLP) models \cite{radford2021learning, li2022blip, li2023blip2, alayrac2022flamingo} have shown significant progress in recent years. A notable advancement is the emergence of zero-shot capabilities, which turn VLP models into generalist models, particularly when combined with large language models (LLMs). These models are now capable of solving various vision-language (VL) downstream tasks, including visual question answering (VQA) and image captioning, without the need for task-specific fine-tuning. The general knowledge enabling such zero-shot capabilities can be attained through training with massive image-text pair datasets \cite{schuhmann2021laion,schuhmann2022laion_5b,gadre2023datacomp}. 

\begin{figure}[t]
  \centering
  \includegraphics[width=\linewidth]{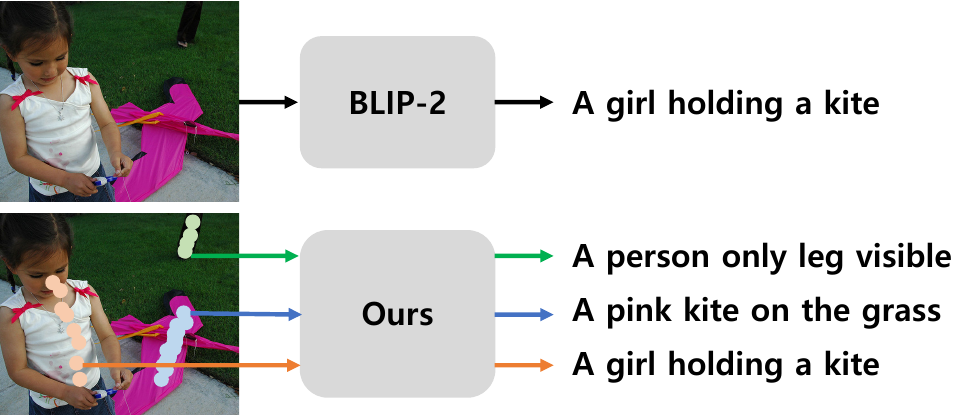} 
  \caption{\label{figure_1} Conceptual comparison between BLIP-2 and our model. While BLIP-2 generates a single caption based on the entire image, our model can generate multiple captions corresponding to regions explicitly indicated by users.}
  \vspace{-0.6em}
\end{figure}

However, VLP models still face a significant challenge: their limited ability to comprehend the fine-grained semantics of specific regions within an image. This stems from the nature of their training datasets. Existing image-text pairs, typically obtained by web crawling, tend to focus on the salient information of the image and fail to provide an explicit indication of the area of the image the text is describing. 
As a result, existing VLP models tend to focus on the implicit global information of the image, lacking the ability to understand the image region explicitly indicated by a user.

\begin{figure*}[t]
  \centering
  \includegraphics[width=0.8\linewidth]{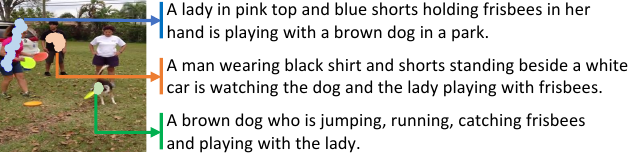} 
  \caption{\label{LN_examples} Examples of trajectories and their corresponding captions provided by the Localized Narratives dataset.}
\end{figure*}

In this paper, we introduce RegionVLM, equipped with the \textbf{regional understanding} capability with explicit, or interactive, indications from users. We argue its significance for the following reasons. 
First, the regional understanding ability broadens the versatility of VLP models. This facilitates the execution of additional vision-language tasks that require explicit indications of regions, such as referring image segmentation \cite{yu2016modeling} and visual commonsense reasoning \cite{zellers2019recognition}.
Second, we can tackle inherent ambiguity (or multiplicity) in VL tasks~\cite{gao2022pyramidclip, chun2022eccv_caption, chun2023improved, chun2021probabilistic} by employing regional understanding. An image can inherently be described by numerous text descriptions, \eg, describing the visual attribute of the salient object, explaining the background, etc. 
However, current VLP models are trained to associate an image with a single given caption rather than encompassing all possible explanations. By empowering a model to focus on specific regions with explicit indications, we can better manage this ambiguity.
Third, integrating regional understanding enhances the interactivity between the model and users. As demonstrated by commercial services such as GPT-4V~\cite{gpt4v, yang2023dawn}, enabling users to specify regions of interest in an image can lead to more precise and relevant interactions.


{There have been several attempts to make VLP models endow region understanding ability. One possible direction is to develop a specialized model based on CLIP \cite{radford2021learning} for segmentation and detection by directly using fully supervised labels for each task \cite{zhou2023zegclip, gu2021open, yun2023ifseg, li2023tagclip}. Several recent studies aim to develop a generalist model with the capability of region understanding by leveraging datasets containing image regions and their corresponding captions \cite{zhang2023gpt4roi, wang2023visionllm, jin2023grill, zhou2023regionblip}. However, the datasets used for these methods exhibit inherent drawbacks. For example, the visual grounding dataset \cite{yu2016modeling} contains region-text pairs only for a limited set of object classes; the captions of Visual Genome \cite{krishna2017visual} are relatively short and only depict a limited relationship between objects. To achieve generality and scalability in VLP models, we need a dataset containing diverse regions with various open-world objects as well as expressive captions.}

{In the paper, we propose to exploit regional textual information from diverse narratives of images. Specifically, we utilize the Localized Narratives (LN) datasets \cite{pont2020connecting, voigtlaender2023connecting}, which provide narrative descriptions by annotators and their mouse trajectory over the described region. The LN dataset includes expressive free-form captions depicting multiple open-world objects in a single image (see Figure \ref{LN_examples}), and thus, it can provide general and meaningful regional information to the VLP model.}
As shown in Figure~\ref{figure_1}, unlike BLIP-2 which can generate captions only for the entire image, our model can generate captions for multiple regions of the image through explicit indications.

We introduce a simple technique allowing a model to accept the regional information of LN. More specifically, we directly convert the 2D coordinates of the trajectory points to the sequence of strings (\eg, ``['', ``19'', ``44'', ``]'', ``['', ``23'', ``55'', ``]'', as shown in Figure \ref{fig_att_samples}) and simply use them as the input of VLP models. Finally, our model is trained to generate a caption corresponding to image regions associated with each trajectory, resulting in an ability to understand regional information. 
Our approach does not require architectural modifications or redefinition of the objective function, ensuring seamless alignment with the original scheme that takes the entire image and text as input.

Our RegionVLM can incorporate various appealing aspects into the existing model while preserving its original capabilities. Our experiments demonstrate that our generalist model can achieve the interactive dialogue system by understanding the explicit region indication from a user. In addition, we show that our model can perform various zero-shot regional understanding tasks that were beyond the capability of the conventional BLIP-2. Furthermore, our model achieves better performance than the recent state-of-the-art methods.

\section{Related Works}\label{relatedworks}
\subsection{Vision-Language Pretraining} 
Vision-language pre-training (VLP) aims to learn meaningful multi-modal representations, enabling zero-shot ability and few-shot adaptation for various VL tasks.
CLIP~\citep{radford2021learning} and its variants~\citep{li2021supervision, mu2022slip, geng2023hiclip} align the vision and language representations obtained from independent vision and language encoders. 
The unified architecture, which learns multi-modal joint representation, is also popularly adopted~\cite{li2022blip, li2021align, chen2020uniter, wang2023image, kim2021vilt} and shows powerful performance on various vision-language tasks.
Recently, the attempts to inject visual information into large language models (LLMs) have been proposed \cite{koh2023grounding, tsimpoukelli2021multimodal, li2023blip2, alayrac2022flamingo}.
They can fully exploit the generality power of LLMs so that they have zero-shot, few-shot adaptation, and in-context learning abilities.
All the models mentioned above are trained only on image-text pairs, so they primarily concentrate on global image information, with a limited understanding of the local regions of the image.

\subsection{Region Modeling for VLP}
To equip VLP models with region-specific information, a dataset explicitly matching image regions to their corresponding texts is essential. 
However, due to the lack of publicly available datasets and the high costs of creating such datasets, researchers often rely on various forms of supervision, though these methods have their limitations. 
For example, datasets which provide object bounding boxes or masks annotated with their class names, such as MS-COCO~\citep{lin2014microsoft,chen2015microsoft} and OpenImages~\citep{kuznetsova2020open}, have been widely utilized~\citep{li2022grounded, wang2023visionllm, zang2023contextual, zhong2022regionclip, zhang2022glipv2}. 
However, the text descriptions in the datasets are short and simple object class names, which have limited ability to capture the relationships between objects in an image.
Visual Genome~\citep{krishna2017visual} provides dense captions of various objects and attributes in an image. Still, its captions are relatively short and simple, falling short in modeling the complex inter-object relationships.
The visual grounding datasets, such as RefCOCO~\citep{yu2016modeling} or visual common-sense reasoning (VCR) dataset~\citep{zellers2019recognition}, have also been utilized~\citep{lai2023lisa, yao2022pevl,zhang2023gpt4roi}.
However, their region-text pairs still provide limited contexts (\eg, 80 class categories for RefCOCO, and person-centric categories for VCR).
We utilize the Localized Narratives dataset~\cite{pont2020connecting, voigtlaender2023connecting}, a comprehensive large-scale dataset that includes expressive captions corresponding to various regions associated with open-world objects.

Image-level prompting is another line of research for enabling regional understanding of VLP models.
\citet{wang2023caption} crop the target image region and feed the cropped image into the model.
\citet{shtedritski2023does} propose drawing a red circle around the target object in the image, which can direct the model's attention to a specific region. 
These methods can provide regional information to VLP models. However, they involve manual manipulation of the original image, potentially contaminating or eliminating the crucial context around the target object.
In contrast, our method does not interrupt the image, preserving all contexts.

\begin{figure*}[t]
  \centering
  \includegraphics[width=\linewidth]{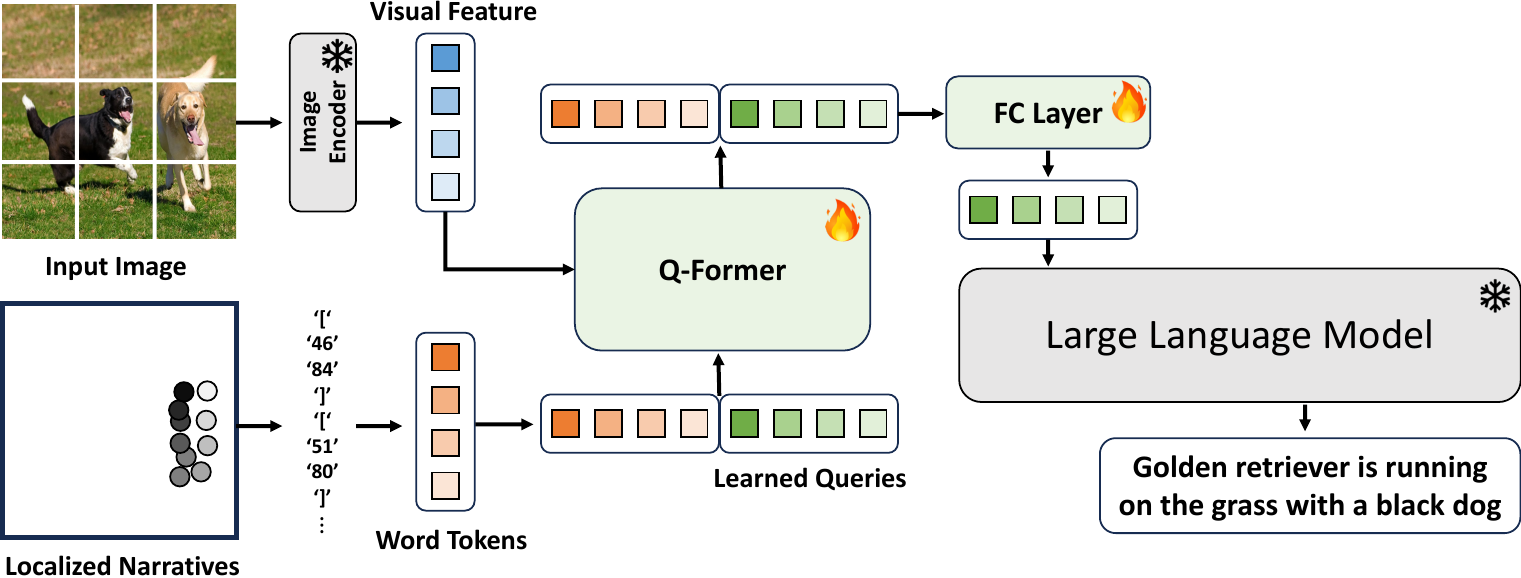} 
  \caption{\label{fig_overview} Overall architecture of our proposed model. Our model converts a set of trajectory points from Localized Narratives into word tokens. The word tokens and visual features are passed to the Q-Former, generating a soft prompt. This allows the frozen LLM to generate captions corresponding to the indicated regions.}
\end{figure*}

\section{Proposed Method}

In this section, we describe our training dataset and the proposed model, which addresses the limitations of previous methods introduced in Section~\ref{relatedworks}.
We first revisit our base model, BLIP-2~\cite{li2023blip2} in Section~\ref{revisitblip2}.
We then introduce our dataset and model in Sections~\ref{method_data} and \ref{method_model}, respectively.
Finally, we present how our method performs various VL downstream tasks in Section~\ref{method_down}.

\subsection{Revisiting BLIP-2}\label{revisitblip2}
In this paper, we use BLIP-2~\citep{li2023blip2} as our base model due to its training efficiency and versatility.
BLIP-2 aims to bridge a frozen pre-trained visual encoder and a frozen large language model (LLM) through a Q-former module, which effectively allows the LLM to comprehend images while maintaining its overall versatility.
Given the input image, the frozen pre-trained visual encoder produces the image feature $I$. 
The Q-former module introduces $N$ learnable input queries $Z$.
These input queries are subsequently updated by interacting with each other
through self-attention layers and interacting with the image
features $I$ through cross-attention layers. After a linear projection, output embedding $\hat{Z} \in \mathbb{R}^{N \times d}$ is obtained: $\hat{Z} = \texttt{Linear}(\texttt{Q-Former}(Z; I))$, where $d$ is the dimension of the text embedding of the LLM. 
Given $\hat{Z}$ to the LLM, the Q-former is trained so that the frozen LLM generates the caption of the given image through language modeling loss. 
The model is trained with image-text pair datasets such as MS-COCO and LAION.

\subsection{Dataset Construction}\label{method_data}
To achieve a generalist model that has zero-shot capabilities with region understanding ability, it is essential to have a dataset containing diverse regions indicating various open-world objects and expressive captions. We explore a new dataset in terms of VLP, the Localized Narratives dataset~\citep{pont2020connecting, voigtlaender2023connecting}.
This dataset includes images accompanied by narrative descriptions from annotators, along with their mouse trajectories over the corresponding regions.
For a single image, an annotator describes the various situations and relationships among objects in the image using several sentences (see Figure~\ref{LN_examples}).
Therefore, we can split the given caption into multiple sentences based on periods (.) and commas (,) and associate each sentence with the corresponding trajectory points.
From now on, in this context, we will refer to the trajectory points as \emph{scribbles}.
The Localized Narratives dataset is large-scale, contains expressive free-form captions depicting open-world objects, and provides multiple scribble-caption pairs for a single image. These properties enable the model to learn general region-aware multi-modal representation.

\begin{figure*}[t]
  \centering
  \includegraphics[width=0.93\linewidth]{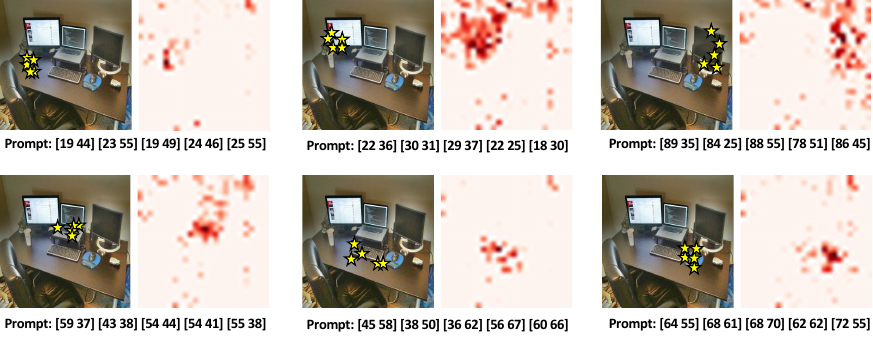} 
  \caption{\label{fig_att_samples} Examples of cross-attention maps between learnable queries $Z$ and image features $I$ by varying the $W$ for a single image. The examples demonstrate that the queries successfully attend to the regions indicated by $W$, as denoted by yellow stars.}
\end{figure*}

\subsection{Grounding Image Regions to LLM}\label{method_model}
We propose a simple yet intuitive technique to convey regional information obtained from Localized Narratives to the frozen LLM. The overall architecture of our model is presented in Figure~\ref{fig_overview}.
Suppose an image with multiple scribbles $\{S\}$ and their associated captions $\{T\}$.
We randomly choose one of the scribble-caption pairs, namely $\{S_i, T_i\}$. From $S_i$, we randomly sample $K$ points, which can be represented as a 2-dimensional list $P = [[x_1, y_1], [x_2, y_2], \cdots , [x_K, y_K]]$, where $x$ and $y$ indicate relative positions of the image (\textit{i.e.,} $0 \leq x, y \leq 1$).

To inject the regional information into the model, we convert $P$ into text.
However, directly using the 2-dimensional list can result in unnecessarily long input tokens. We introduce two tricks to simplify the text string.
First, we removed unnecessary redundant word tokens, including those for the outermost brackets and intermediate commas.
Second, we multiplied each coordinate by 100 and rounded them to ensure they have integer values. This allows us to omit the repeated ``0.'' string.
For example, when $K=2$, $P=[[0.324$, $0.643]$, $[0.369$, $0.622]]$ is converted to a string ``\texttt{[32} \texttt{64]} \texttt{[37} \texttt{62]}''. 
We then tokenize the string by using a tokenizer, resulting in a set of word tokens $W \in \mathbb{R}^{L \times d}$, where $L$ is the length of word tokens in $W$.

In the Q-former module, the learnable input queries $Z$ are concatenated with the word tokens $W$. This enables the queries $Z$ to engage in cross-attention mechanisms with the visual feature $I$, while being conditioned by $W$ through self-attention layers.
As a result, the Q-former produces the output query embeddings $\hat{Z}$ and the output word embeddings $\hat{W}$, where $[\hat{Z}, \hat{W}] = \texttt{Linear}(\texttt{Q-Former}([Z, W]; I)) \in \mathbb{R}^{(N+L) \times d}$. 
We expect that the output query embeddings $\hat{Z}$ contain the semantics corresponding to the regions indicated by $W$, as the attention mechanism with $W$ enables $Z$ to direct its focus toward the regions.

We provide an empirical analysis showing that the text-form input $W$ can effectively guide the Q-former queries $Z$ to focus on the regions indicated by $W$.
We investigate which image regions the queries attend to by analyzing the attention scores of a cross-attention layer between queries and image features.
We visualize the cross-attention maps between $Z$ and $I$ by varying $W$ for a single image in Figure \ref{fig_att_samples}. It demonstrates that the queries appropriately attend to the regions that the text prompts $W$ actually indicate, which are noted by yellow stars. 

We now provide the output query embeddings $\hat{Z}$ obtained from the Q-former to the frozen LLM. Note that we exclusively use $\hat{Z}$ while excluding $\hat{W}$ to maintain the original scheme of BLIP-2.
We train the Q-Former using the language modeling loss so that the frozen LLM generates the text $T_i$ corresponding to the region associated with $S_i$.
However, focusing mainly on modeling the local region may lead to a loss of global image understanding ability.
Therefore, half of the training mini-batch samples are sampled from the Localized Narrative dataset, while the remaining half is sourced from global image-text pairs originating from the existing dataset (\eg, LAION), following the standard training procedure of BLIP-2.
Note that we set $S$ as an empty string for the global image-text pairs, denoted by ``\,''.

\begin{figure*}[t!]
  \centering
  \includegraphics[width=1.05\linewidth]{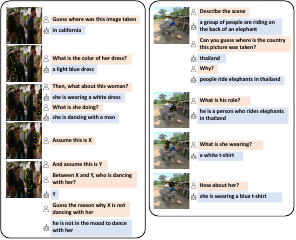} 
  \caption{\label{fig_demo} Selected examples of interactive dialogue using our model. The regions indicated by a user are noted as yellow stars. The examples illustrate a wide range of abilities for interacting with users, reasoning, guessing, question answering, etc. Note that the series of dialogues in one column is obtained from a single process.}
\end{figure*}

\subsection{Downstream Vision-Language Tasks}\label{method_down}
Our generalist model can be utilized in a range of VL downstream tasks that demand either global or local image comprehension abilities, or both.

\textbf{Visual Question Answering (VQA):} It requires the ability for global image understanding.
Given an image, the Q-former generates the output query embeddings $\hat{Z}$ using $S$=``\,'', ensuring that $\hat{Z}$ contains the global image information.
We attach the word tokens of a question text prompt (``Question: \{\} Answer:'') following $\hat{Z}$ as input to the LLM. 
We follow BLIP-2~\citep{li2023blip2} for the answer generation process.

\textbf{Referring Image Segmentation (RIS):} It aims to segment the object based on the provided language description.
Zero-shot RIS can directly showcase the model's region modeling ability. 
To implement zero-shot RIS, we first obtain several object proposals $\{M_i\}$ using the Segment Anything Model (SAM)~\citep{kirillov2023segment}. Among these proposal masks, our goal is to select a mask whose caption, generated by our model, is mostly similar to the given language description $Y$.
We generate $K$ random points inside each $M_i$ and convert these points into the text prompt $W_i$.
The $W_i$ is then fed into our trained model along with the image features $I$, resulting in the likelihood of the generated text $y_i$. We compare each $y_i$ with $Y$ and select the final output mask $M_{i^*}$, where $i^* = \argmin_i \texttt{dist}(y_i, Y)$, and \texttt{dist} is a distance metric. Since we have no access to supervision for RIS, there is little chance that the generated captions closely resemble $Y$ as provided by the RIS datasets. Therefore, we define \texttt{dist} as the language modeling loss. 
We can expect improved performance by exploring more advanced distance metrics, but this is beyond our current scope.

\begin{table*}[tbp]
  \centering
  \caption{Comparison with recent state-of-the-art weakly supervised and zero-shot methods for referring image segmentation on three benchmarks. Our results are obtained by a single experiment run.}

    \begin{tabular}{lcccccccc}
     \Xhline{1pt}\\[-0.95em]
               &        \multicolumn{3}{c}{RefCOCO} & \multicolumn{3}{c}{RefCOCO+} & \multicolumn{1}{c}{GRef} \\
    Method   &\textit{val}  & \textit{testA} & \textit{testB}  &\textit{val}  & \textit{testA} & \textit{testB} &\textit{val}  \\
    \hline\hline \\[-0.9em]
   
    \multicolumn{3}{l}{Supervision: Weakly Supervised} \\
    \text{TSEG}~\citep{strudel2022weakly} & 25.44& -&-& 22.01  & - & - &22.05 \\
 
        \text{\citet{liu2023referring}} &  31.17 &32.43 &29.56 &30.90 &30.42 &30.80 &36.00  \\
    \text{\citet{kim2023shatter}}& 34.76 &34.58& 35.01 &28.48& 28.60 &27.98 &28.87\\
    \text{\citet{lee2023weakly}}& 31.06 &32.30& 30.11 &31.28& 32.11 &30.13 &32.88\\
            \\[-0.9em]
\hline
    \\[-0.9em]
    \multicolumn{3}{l}{Supervision: Zero-Shot} \\
    \citet{yu2023zero} &  26.70 & 24.99 & 26.48 & 28.22 & 27.54 & 27.86 & 32.79  \\

    RegionVLM (Ours)  & \textbf{38.74} & \textbf{39.40} & \textbf{37.59}&\textbf{31.47} &\textbf{33.99} & \textbf{30.22}&\textbf{33.94}\\
 
    \Xhline{1pt}
    \end{tabular}%
  \label{table_RIS}%
\end{table*}%
\begin{figure*}[t]
  \centering
  \includegraphics[width=\linewidth]{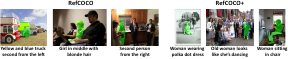} \\[-0.5em]
  \caption{\label{fig_ris} Selected examples of referring image segmentation on RefCOCO (\textit{left}) and RefCOCO+ (\textit{right}).}
  \vspace{-0.6em}
\end{figure*}

\textbf{Visual Commonsense Reasoning (VCR):} Given a set of object masks in the image, VCR aims to answer the questions related to those objects (Figure~\ref{fig_VCR}).
Moreover, beyond merely answering about the question, the model is required to provide a justification of the generated answer.
Therefore, VCR demands advanced cognition and commonsense reasoning abilities, as well as both global and local image understanding abilities.
Since VCR requires the information for multiple objects, we implement VCR as follows: for each given object mask $M_i$, we generate $K$ random points inside it and obtain $\hat{Z}_i$ through the Q-Former.
We then create a prompt to let the LLM understand which object index is associated with each query embedding, and we append the question to this prompt.
The resulting example input to the LLM is as follows: ``[0]: $\hat{Z}_0$ [1]: $\hat{Z}_1$, What is [0] here to do? 1. [1] is here to steal gold from [0]. 2. $\cdots$ 3. $\cdots$ 4. $\cdots$''.
Our model will respond with one of choices 1 through 4, which it considers the most appropriate.

\section{Experiments}
\subsection{Experimental Setup}\label{section_setup}
Our base model is pre-trained BLIP-2~\citep{li2023blip2} equipped with FlanT5$_\text{XL}$~\citep{chung2022scaling}. For a visual encoder, we use ViT-g/14 from EVA-CLIP~\cite{fang2023eva}.
We finetune the Q-former and the linear layer of BLIP-2 for 10 epochs with a learning rate of $5 \times 10^{-6}$ and a batch size of 64. We set $K$ to 10 and $N$ to 32. We follow the configuration of BLIP-2 for other settings regarding optimization.
For experiments, we use 8 NVIDIA Tesla V100 (32GB) GPUs.
For the training dataset, we use a mixture of Localized Narratives datasets built upon images \cite{pont2020connecting} and videos \cite{voigtlaender2023connecting}.
Additionally, we utilize the Visual Genome dataset~\cite{krishna2017visual} that provides bounding boxes along with their captions. To generate the set of points $P$ for this dataset, we sample random $K$ points inside the bounding box. For global image-text pairs, we use 115M images from the LAION-400M dataset~\cite{schuhmann2021laion} filtered by \citet{li2022blip}.

\begin{table}[t]
  \centering
  \caption{Comparison with recent state-of-the-art methods for zero-shot visual commonsense reasoning. Our results are obtained by a single experiment run.}
    \resizebox{0.49\textwidth}{!}{
    \begin{tabular}{@{\hskip 0.03in}lccccc@{\hskip 0.03in}}
    \Xhline{1pt}\\[-0.95em]
           & Q$\rightarrow$A   & QA$\rightarrow$R  & Q$\rightarrow$AR \\
    \hline\hline \\[-0.9em]
    Random  & 25.0 & 25.0 & 6.3 \\
    VL-T5~\citep{cho2021unifying}  & 28.2 &27.5 & 8.2 \\
    FewVLM~\citep{jin2021good} &27.0 & 26.1 & 7.4\\
    GRILL~\citep{jin2023grill} &40.6 & 39.3 & 16.2 \\
    UniFine~\citep{sun2023unifine} & \textbf{58.3} & 51.3 & - \\
    RegionVLM (Ours) &   52.4  & \textbf{54.6}      & \textbf{29.3} \\

    \Xhline{1pt}
    \end{tabular}%
    }
  \label{table_vcr}%
\end{table}%

\subsection{Experimental Results}

\textbf{Interactive Dialogue System:} 
Our model can enable an interactive dialogue system with the generality power from the frozen LLM, realized by appending the previous chat history in front of the new query.
If the image is provided, we append the Q-former queries computed from the image with user region indication in front of the text prompts.
In Figure~\ref{fig_demo}, we provide examples illustrating its capacity to comprehend the region indicated by the user (interactivity) and its ability for reasoning, guessing, and answering questions.
It's worth noting that our method also retains the original BLIP-2's capability to process and understand the entire image.

\begin{figure*}[t]
  \centering
  \includegraphics[width=\linewidth]{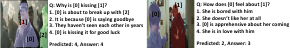} \\[-0.5em]
  \caption{\label{fig_VCR} Selected examples of visual commonsense reasoning.}

\end{figure*}

\begin{table}[t]
  \centering
  \caption{Comparison with BLIP-2~\citep{li2023blip2} on zero-shot visual question answering.}
    \begin{tabular}{l@{\hskip 0.05in}c@{\hskip 0.1in}c@{\hskip 0.1in}c}
    \Xhline{1pt}\\[-0.95em]
          & OK-VQA & GQA   & VQAv2 \\
    \hline\hline \\[-0.9em]
    BLIP-2  &   41.08    &   \textbf{43.92}    & 63.12 \\
    RegionVLM (Ours)  &    \textbf{41.88}   &   43.50    &  \textbf{63.22} \\
    \Xhline{1pt}
    \end{tabular}%
  \label{table_zeroshotvqa}%
\end{table}%

\textbf{Zero-shot RIS:} 
Table~\ref{table_RIS} compares our method with recent state-of-the-art weakly supervised and zero-shot RIS methods on three benchmarks: RefCOCO~\cite{yu2016modeling}, RefCOCO+~\cite{yu2016modeling}, and G-Ref~\cite{mao2016generation}. We adopt mean Intersection-over-Union (mIoU) as an evaluation metric. As shown in Table~\ref{table_RIS}, our generalist method achieves significantly better performance than \citet{yu2023zero}, which is a current state-of-the-art specialized zero-shot RIS method. 
Additionally, our method demonstrates competitive performance compared to recent weakly supervised RIS methods~\cite{strudel2022weakly, liu2023referring, kim2023shatter, lee2023weakly}, which have access to image-text pairs for each target dataset.
Figure~\ref{fig_ris} presents examples of RIS results obtained by our model.

\textbf{Zero-shot VCR:}
Table~\ref{table_vcr} compares our method with recent state-of-the-art zero-shot visual commonsense reasoning on the VCR dataset~\cite{zellers2019recognition}.
The benchmark comprises three evaluation tasks: question answering (Q $\rightarrow$ A), rationale prediction given a question and answer pair (QA $\rightarrow$ R), and answer and rationale prediction given a question (Q $\rightarrow$ AR). 
We use accuracy as an evaluation metric. As shown in Table~\ref{table_vcr}, our method demonstrates comparative performance compared to the recent zero-shot VCR methods.
It outperforms GRILL~\cite{jin2023grill} by 11.8\%p on Q $\rightarrow$ A, 15.3\%p on QA $\rightarrow$ R, which suggests that our model exhibits particularly strong reasoning ability.
UniFine~\cite{sun2023unifine} shows superior Q $\rightarrow$ A performance compared to ours, whereas our model outperforms UniFine in QA $\rightarrow$ R, also indicating our stronger reasoning abilities. 
Our method focuses on generative modeling with a frozen LLM, known for its strong reasoning ability. On the other hand, UniFine~\cite{sun2023unifine} utilizes frozen CLIP~\cite{radford2021learning}, Roberta~\cite{liu2019roberta}, and OFA~\cite{wang2022ofa}, which may exhibit weaker reasoning capabilities compared to recent strong LLMs.
Additionally, our fine-grained modeling captures meaningful relationships between objects, thereby contributing to enhanced reasoning capabilities.
Figure~\ref{fig_VCR} presents examples of VCR results obtained by our model. Our model generally possesses a commonsense reasoning ability but tends to struggle with reasoning about open-world knowledge (\eg, the relationship between Harry Potter and Dudley's mom, as shown in the right example).


\begin{table}[t]
  \centering
  \caption{Comparison of BLIP-2~\citep{li2023blip2} combined with various region modeling methods for zero-shot referring image segmentation on the RefCOCO, RefCOCO+, and G-Ref validation sets.}
\resizebox{0.49\textwidth}{!}{
    \begin{tabular}{@{\hskip 0.03in}l@{\hskip 0.08in}c@{\hskip 0.08in}c@{\hskip 0.08in}c@{\hskip 0.03in}}
    \Xhline{1pt}\\[-0.95em]
          & RefCOCO & RefCOCO+   & GRef \\
    \hline\hline \\[-0.9em]
    \citet{shtedritski2023does}  &    14.85    & 15.76 & 15.86  \\

    \citet{wang2023caption}  &   27.91    &   31.45    & 31.38 \\
    RegionVLM (Ours)  &    \textbf{38.74}   &   \textbf{31.47}    &  \textbf{33.94} \\
    \Xhline{1pt}
    \end{tabular}%
}
  \label{table_blipvariant}%
\end{table}%

\textbf{Zero-shot VQA:} We conduct a quantitative assessment for zero-shot VQA on OK-VQA~\cite{marino2019ok}, GQA~\cite{hudson2019gqa}, and VQAv2~\cite{goyal2017making} benchmarks. 
Table~\ref{table_zeroshotvqa} demonstrates that our method achieves comparable performance with BLIP-2, suggesting that our approach preserves the global image understanding ability.

\begin{table}[t]
  \centering
  \caption{Comparison with BLIP-2~\cite{li2023blip2} for zero-shot image captioning on the NoCaps dataset.}
    \begin{tabular}{l@{\hskip 0.1in}c@{\hskip 0.1in}c@{\hskip 0.1in}c}
    \Xhline{1pt}\\[-0.95em]
          & BLEU@4 & SPICE   & CIDEr \\
    \hline\hline \\[-0.9em]
    BLIP-2  &  43.4	& 14.0	& 105.8 \\
    RegionVLM (Ours)  &    \textbf{47.7}&	\textbf{15.5}&	\textbf{119.1}
 \\
    \Xhline{1pt}
    \end{tabular}%
  \label{table_zeroshotcap}%
\end{table}%

\textbf{Zero-shot Captioning:}
Table~\ref{table_zeroshotcap} presents the zero-shot captioning performance on the NoCaps~\cite{agrawal2019nocaps} benchmark. Compared to BLIP-2, our model achieved improved performance across all three evaluation metrics: BLEU@4, SPICE, and CIDEr. We believe that our regional modeling contributes to the model's ability to capture fine-grained information, resulting in descriptive and detailed captions.

\textbf{Comparison with other region modeling methods:} We compare our method with two recent techniques that can inject regional information into BLIP-2 on RIS. 
We used the same evaluation settings, including the mask proposals from SAM~\cite{kirillov2023segment} and the matching process between the generated captions and give descriptions.
For \citet{shtedritski2023does}, the image with a red circle drawn on each proposal area was inserted into BLIP-2 to generate a caption.
For \citet{wang2023caption}, the cropped box corresponding to each proposal is resized to the original image size, and BLIP-2 generates the caption based it.
Table~\ref{table_blipvariant} demonstrates that our method achieves significantly better performance compared to those two methods, and comparable performance with \citet{wang2023caption} on RefCOCO+.  
The language descriptions from RefCOCO+ tend to depict the object itself with less focus on its surrounding contexts or location.
However, RefCOCO depicts the surrounding context of the object such as its location, so global information should be considered together (see Figure~\ref{fig_ris}). 
\citet{wang2023caption} inject only region-of-interest into the model by cropping the proposal region, thereby losing the proposal's contexts and location. Our method can consider both local and global information for the proposal, yielding satisfactory results across all benchmarks for RIS.

\begin{table}[t]
  \centering
  \caption{Robustness of our model against the noisy input scribbles.}
  \resizebox{0.42\textwidth}{!}{
    \begin{tabular}{lcccc}
    \Xhline{1pt}\\[-0.95em]
          Dilation & 0 & 3   & 7 & 15 \\
    \hline\hline \\[-0.9em]
    mIoU & 37.73&	37.64	&36.39	&35.77\\

    \Xhline{1pt}
    \end{tabular}%
    }

  \label{tab_robust}%
\end{table}%

\subsection{Discussion}
\textbf{Robustness against the noisy input scribbles:}
Our interactive system expects user scribble inputs. However, in practice, scribbles obtained from users can be fall outside the intended object.
We argue that our model is robust against noisy user input because the scribbles in the Localized Narratives dataset are collected through the free-form mouse movements of human annotators, which are inherently noisy.
We support this argument with an additional quantitative analysis on RIS. As described in Section~\ref{method_down}, the object proposal masks are obtained by SAM~\cite{kirillov2023segment}. Instead of utilizing the SAM-generated mask directly, we introduce some noise to simulate a noisy scribble environment. More precisely, we enlarge each SAM-generated mask by varying dilation ratios. This simulation represents a scenario in which a user provides a coarse mouse scribble that is not perfectly aligned with the target object but could contain the outside of the object. Table~\ref{tab_robust} demonstrates the robustness of our model against the noisy input scribbles.

\begin{table}[t]
  \centering
  \caption{Comparison of referring image segmentation performance by varying $K$.}
  \resizebox{0.42\textwidth}{!}{
    \begin{tabular}{lccc}
    \Xhline{1pt}\\[-0.95em]
          & RefCOCO & RefCOCO+   & GRef \\
    \hline\hline \\[-0.9em]
    $K=5$  &  37.56     & 30.95 &  32.75\\

    $K=10$  &   \textbf{38.74}   &   \textbf{31.47}    &  \textbf{33.94} \\
    $K=15$  & 37.98     &   31.14    &  33.11 \\
    \Xhline{1pt}
    \end{tabular}%
    }

  \label{table_sensitivity}%
\end{table}%

\textbf{Sensitivity to $K$:} We analyze the sensitivity of the RIS performance to the value of $K$. Table~\ref{table_sensitivity} presents the RIS performance, varying $K$ at test time, using the model trained with $K=10$.
This demonstrates that our model operates successfully even when the number of provided points differs between training and testing.

\begin{table}[t]
  \centering
  \caption{Effectiveness of point representation design using Localized Narratives (LN) on zero-shot VQA.}
    \begin{tabular}{l@{\hskip 0.2in}c@{\hskip 0.15in}c@{\hskip 0.15in}c}
    \Xhline{1pt}\\[-0.95em]
          & OK-VQA & GQA   & VQAv2 \\
    \hline\hline \\[-0.9em]
    LN w/o points  &   	39.56	& 42.90&	61.74 \\
    LN w/ points  &    \textbf{41.88}   &   \textbf{43.50}    &  \textbf{63.22} \\
    \Xhline{1pt}
    \end{tabular}%
  \label{table_ln_naive}%
\end{table}%

\textbf{Leveraging Localized Narratives without Point Representation:}
We can study the effectiveness of our proposed method by training the model trained with LN without the point representation. However, without the point representation, the model cannot perform tasks that require explicit region indication. Therefore, we present zero-shot performance on VQA for the model fine-tuned by using LN + LAION without the point representation. Table~\ref{table_ln_naive} shows that the point representation brings better VQA performance. 
We believe this is because LN contains less diverse and descriptive captions compared to the existing image-text pair datasets.
However, our proposed model can preserve global understanding ability by separating the learning of global understanding and local understanding through the point representation.

\section{Conclusion}
In this study, we have addressed the limited region understanding ability of existing vision-language pre-training models.
We proposed a model that can input the indication of the region, which isseamlessly integrated into the existing model. 
In addition, we utilized Localized Narratives to learn the general knowledge of image regions. 
Our experiments showcase the superior performance of our generalist model across a diverse set of zero-shot region understanding tasks, without compromising its ability for global image comprehension tasks. As a generalist model, we foresee significant potential for further enhancement through instruction tuning~\cite{liu2023visual, instructblip}, establishing a promising direction for future research.

\bibliography{anthology,custom}

\clearpage

\appendix

\section{Appendix}
\label{sec:appendix}

\paragraph{Dataset Details.}
As mentioned in Section~\ref{section_setup}, we use a combination of image Localized Narratives~\cite{pont2020connecting}, video Localized Narratives~\cite{voigtlaender2023connecting}, Visual Genome~\cite{krishna2017visual}, and LAION-400M filtered by \citet{li2023blip2}. We present statistics of each dataset in Table~\ref{table_stat}.

\begin{figure}[t]
  \centering
  \includegraphics[width=\linewidth]{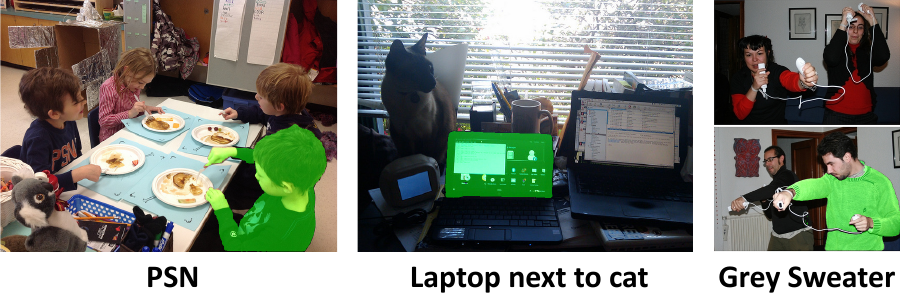} \\[-0.5em]
  \caption{\label{figure_fail} Examples of our failure cases on zero-shot referring image segmentation.}
  \vspace{-0.6em}
\end{figure}
\paragraph{Failure Case Analysis for RIS.}
As demonstrated in Table~\ref{table_RIS} and Figure~\ref{fig_ris}, our method successfully performs zero-shot RIS, but may occasionally yield unsatisfactory results. Figure~\ref{figure_fail} shows some failure cases.
Figure~\ref{figure_fail}(\textit{left}) shows that our model tends to struggle to recognize the character in the image. We believe that we can further utilize Optical Character Recognition (OCR)~\cite{baek2019character, baek2019wrong, kim2022ocr} datasets, which provide characters in an image together with their location. 
Figure~\ref{figure_fail}(\textit{middle}) shows that our method correctly identifies the target object, but tends to focus on small partial regions of the target object. 
This limitation is also explored in recent weakly supervised segmentation studies~\cite{lee2021reducing, lee2021anti}.
We conjecture that the reason for this is that only small regions of the target object can provide sufficient information to generate captions that align with the given language descriptions.
Figure~\ref{figure_fail}(\textit{right}) shows that our method produces lower accuracy of RIS although our method identified the referred object successfully. The language descriptions from RIS datasets tend to describe a person by using only a portion of the individual, such as ``grey sweater'' in the example. Therefore, our method successfully identifies the ``grey sweater'' only, but since the actual ground truth includes all regions of a man wearing the grey sweater, these cases impact the overall performance.

\paragraph{Limitations.}
In contrast to global image-text pair datasets, which can be automatically collected from the web, our dataset may have limitations in terms of scalability.
To address this, we can generate pseudo region captions using our trained model, as BLIP-2 utilizes the pseudo captions produced by BLIP~\cite{li2022blip} captioning model.
Additionally, our current evaluations focus mainly on zero-shot downstream tasks. It is also worth to explore the possibility of our method for transfer learning~\cite{yoo2023improving}, semi-supervised learning~\cite{lee2019ficklenet, lee2022anti, lee2022weakly}, few-shot learning~\cite{jin2023grill, alayrac2022flamingo}, and weakly supervised learning~\cite{lee2023weakly, lee2021bbam}.

\paragraph{Potential Risks.}
Since our model is based on frozen LLMs, it shares similar potential risks to LLMs, such as generating offensive output, vulnerability to attacks, and leaking personal sensitive data. To address this, we may need an additional filtering module to prevent such output from being conveyed to users.

\begin{table}[t]
  \centering
  \caption{Number of images and number of region-caption pairs for each dataset.}
    \resizebox{0.49\textwidth}{!}{

    \begin{tabular}{lcc}
    \Xhline{1pt}\\[-0.95em]
          & \# of images & \# of region-caption pairs \\
    \hline\hline \\[-0.9em]
    Image LN  &  306K     & 445K \\

    Video LN  &   125K   &   149K    \\
    Visual Genome  & 77K     &   1.7M   \\
    LAION  & 115M    &   115M   \\

    \Xhline{1pt}
    \end{tabular}%
    }

  \label{table_stat}%
\end{table}%

\end{document}